\pdfoutput=1
\documentclass[11pt]{article}

\usepackage{emnlp2021}

\usepackage{times}
\usepackage{latexsym}

\usepackage[T1]{fontenc}
\usepackage[utf8]{inputenc}

\usepackage{microtype}

\usepackage{amsmath,amsfonts,bm}

\def\eqref#1{equation~\ref{#1}}
\def\1{\bm{1}}

\DeclareMathAlphabet{\mathsfit}{\encodingdefault}{\sfdefault}{m}{sl}
\SetMathAlphabet{\mathsfit}{bold}{\encodingdefault}{\sfdefault}{bx}{n}

\DeclareMathOperator*{\argmax}{arg\,max}
\DeclareMathOperator*{\argmin}{arg\,min}

\usepackage{enumitem}
\usepackage{hyperref}
\usepackage{url}
\usepackage{graphicx}
\usepackage{booktabs}
\usepackage{multirow}

\title{\textsc{FastIF}: Scalable Influence Functions for\\Efficient Model Interpretation and Debugging}

\author{
 Han Guo$^{1,2,3}$ \ \ \ \ \ \ \
 Nazneen Fatema Rajani$^{1}$ \ \ \ \ \ \ \ 
 Peter Hase$^{3}$ \\
 \textbf{Mohit Bansal}$^{3}$ \ \ \ \ \ \ \ 
 \textbf{Caiming Xiong}$^{1}$ \vspace{5pt} \\
 $^{1}$Salesforce Research~~~~~
 $^{2}$Carnegie Mellon University~~~~~
 $^{3}$UNC Chapel Hill \\
 \small\texttt{\{nazneen.rajani, cxiong\}@salesforce.com}\\
 \small\texttt{ hanguo@cs.cmu.edu},
 \small\texttt{ \{peter, mbansal\}@cs.unc.edu}
 }

\begin{document}
\maketitle
\begin{abstract}
Influence functions approximate the ``influences'' of training data-points for test predictions and have a wide variety of applications.
Despite the popularity, their computational cost does not scale well with model and training data size. We present~\textsc{FastIF}, a set of simple modifications to influence functions that significantly improves their run-time. We use $k$-Nearest Neighbors ($k$NN) to narrow the search space down to a subset of good candidate data points, identify the configurations that best balance the speed-quality trade-off in estimating the inverse Hessian-vector product, and introduce a fast parallel variant. Our proposed method achieves about $80$X speedup while being highly correlated with the original influence values.
With the availability of the fast influence functions, we demonstrate their usefulness in four applications.
First, we examine whether influential data-points can ``explain'' test time behavior using the framework of simulatability.
Second, we visualize the influence interactions between training and test data-points.
Third, we show that we can correct model errors by additional fine-tuning on certain influential data-points, improving the accuracy of a trained MultiNLI model by $2.5\%$ on the HANS dataset.
Finally, we experiment with
a similar setup but fine-tuning on data-points not seen during training,
improving the model accuracy by $2.8\%$ and $1.7\%$ on HANS and ANLI datasets respectively.
Overall, our fast influence functions can be efficiently applied to large models and datasets, and our experiments demonstrate the potential of influence functions in model interpretation and correcting model errors.\footnote{
Code is available at \url{https://github.com/salesforce/fast-influence-functions}.
}

\end{abstract}

\section{Introduction}

\begin{figure}
\includegraphics[width=0.99\linewidth]{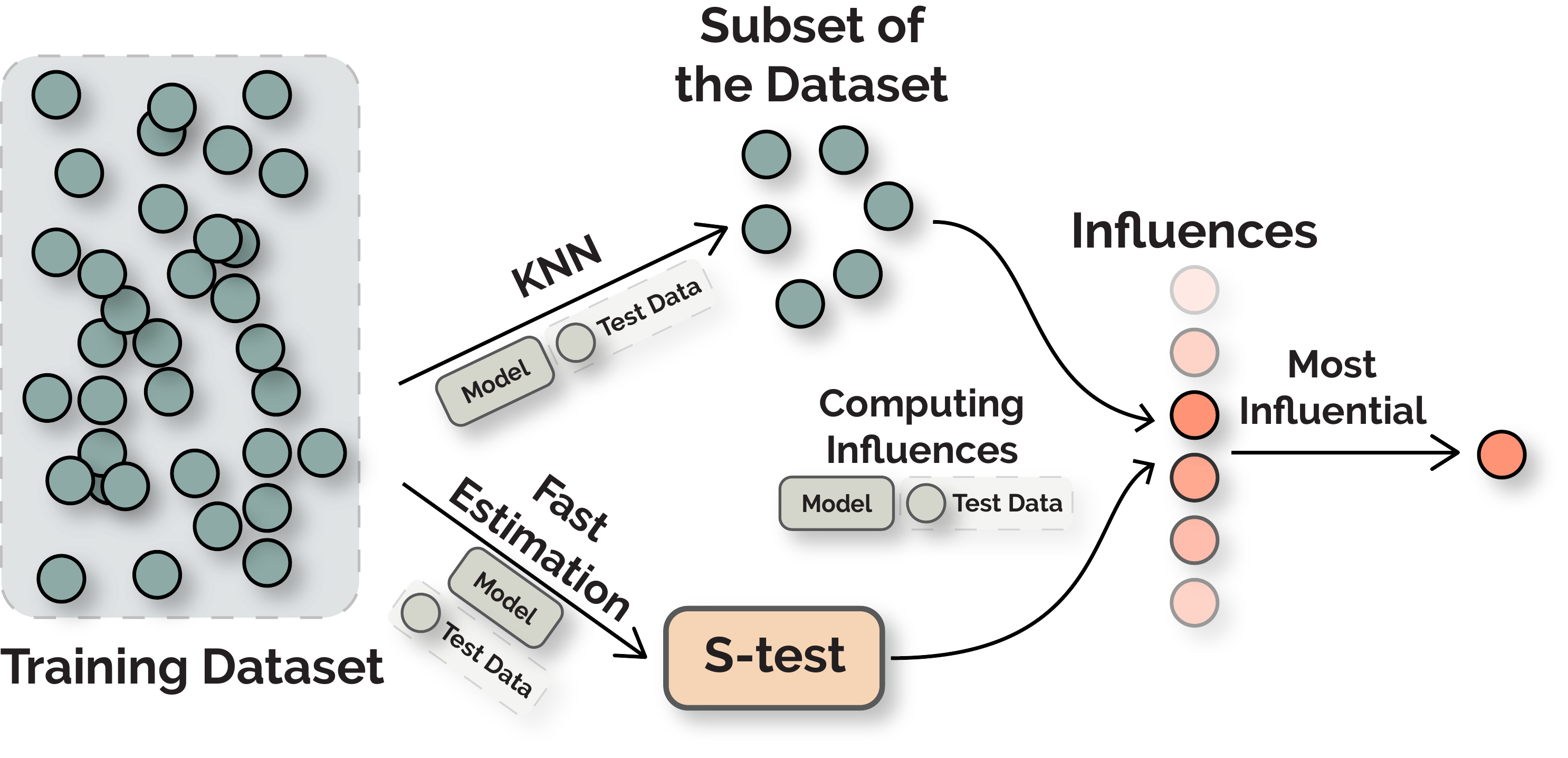}
\centering
\vspace{-10pt}
\caption{Workflow of \textsc{FastIF} w.r.t. a test data-point. 
First a subset of data-points are selected from the entire training set using $k$NN to reduce search space, then the inverse Hessian-vector product ($s_{\text{test}}$) is estimated based on Sec.~\ref{subsec:speedup-ihvp}. The influence values of data-points are computed using the outputs from these two steps. Finally, the most influential data-point(s) are returned.
}
\vspace{-10pt}
\label{figure:main}
\end{figure}

Language understanding systems are becoming ever more powerful with the recent advances in large-scale pre-training. As these systems become widely adopted for real-world applications, the ability to interpret the model decisions grows increasingly important. An array of interpretation methods now exist for shedding light on models' decision-making processes, with the bulk of work focusing on estimating feature importance~\citep{li_visualizing_2016, ribeiro_why_2016,lundberg_unified_2017, sundararajan_axiomatic_2017, ribeiro_anchors_2018} and interpreting internal model representations \citep{donnelly2019interpretability,mu2020compositional,de2020decisions}.
These approaches aim to identify features a model uses to make decisions
or analyze representations obtained from trained models. In contrast, one might also want to know how particular training data points influence model behavior at test time. This kind of goal is an instance of what~\citet{lipton2017} terms ``algorithmic transparency,'' or, transparency ``at the level of the learning algorithm itself.''
The ability to do so would allow researchers to identify and respond to data responsible for problematic test-time behavior.

One simple brute-force way to estimate the importance of a training data-point to a test-time decision is the leave-one-out approach~\cite{hastie2009elements}.
Alternatively, influence functions~\cite{koh2017understanding} provide a tractable estimate of the effect without the need to repeatedly retrain the model.
Yet, influence functions are very expensive to compute for moderate-sized model and training data. For instance, finding the most influential examples w.r.t. an evaluation data-point with a model of about $14.7$M parameters in a dataset of about $390$K examples -- a pretty moderate setting -- takes more than $2$ hours (please see Sec.~\ref{subsec:computation-times} for details). Consequently, applications might face the dilemma of either accepting the computation cost or resorting to a smaller-scale setting.\footnote{For instance, in Footnote~3 of~\citet{han2020explaining} the authors mentioned that they had to resort to using a smaller model and dataset due to the computation cost.} But why are influence functions computationally expensive? In our early experiments, we found that the main computational bottleneck lies in the following steps: 
\begin{enumerate}[leftmargin=*, noitemsep]
\setlength\itemsep{0em}
    \item First, searching for the most positive/negative influential data-point(s) w.r.t. some evaluation data-point(s) is an $\mathcal{O}(n)$ operation (via enumerating the entire training set), and can take more than two hours in our experiments.
    \item Second, it is expensive to estimate the inverse Hessian of model parameters required to compute the influence of a single data-point is expensive (usually in the order of minutes). 
    \item Lastly, previous algorithms perform serial computations that can actually be parallelized. 
\end{enumerate}

In this work, we present~\textsc{FastIF} (Fast Influence Functions) to address these challenges through three simple techniques.
First, instead of enumerating the full training dataset
to find influential data-point, we leverage fast nearest neighbor search~\cite{JDH17} to narrow the search to a small subset of influence-worthy data-points. This operation reduces the computation by about an order of magnitude. Second,
we identify a set of hyperparameters for the Hessian estimation algorithm that reduces computation time by more than half while preserving estimation quality. Finally, we describe a simple parallelizable extension,
which gives an additional $2$X speedup.
As a result, we are able to improve the most influential examples search overall by approximately two orders of magnitude in our experiments. 

So what could we do with faster influence functions? We demonstrate the advantage of fast influence functions via several interesting downstream applications.
These require computing influence functions repeatedly and were thus almost intractable without such fast influence functions:
\begin{enumerate}[leftmargin=*, noitemsep]
\setlength\itemsep{0em}
    \item
    In Sec.~\ref{subsec:explainability-of-influential-examples} we examine the ``explainability'' of influential examples using the framework of simulatability~\cite{doshi2017towards,hase2020evaluating}, and we find that they improve model simulatability.

    \item We visualize how different training data-points interact with different test data-points in Sec.~\ref{subsec:effect-visualization}. The visualizations help us uncover interesting latent structures in the dataset, and the insights could be used to improve model performance.

    \item In Sec.~\ref{subsec:improving-model-prediction} we show a simple application of influence functions in correcting model prediction by fine-tuning on certain influential examples. 
    Despite its simplicity, our method can improve a MultiNLI model's accuracy on the HANS dataset by more than $5.8\%$ (about $2.5\%$ more than our baseline).

    \item Influence functions are defined w.r.t. the training data used in the model training. Also in Sec.~\ref{subsec:improving-model-prediction}, we extend the above experiment and demonstrate the potential of influence functions to the setting where we want to leverage a new labeled dataset unseen during training.
    This can improve a MultiNLI model's accuracy on the HANS/ANLI dataset~\cite{williams2018broad,mccoy2019right,nie2020adversarial} by more than $5.9\%$/$2.7\%$ (about $2.8\%$/$1.7\%$ more than our baseline).
\end{enumerate}

\section{Related Work}

\paragraph{Explainable Machine Learning.}
With the rise of model complexity and adoption of large-scale machine learning models, there is an increasing emphasis on interpreting model behaviors~\cite{doshi2017towards,ribeiro2020beyond,wallace2019allennlp,hase2020evaluating}. 
For example, previous work has considered interpreting predictions by constructing importance scores over the input tokens~\cite{belinkov2019analysis}, attention-based methods~\cite{jain2019attention,wiegreffe2019attention,zhong2019fine}, gradient-based methods~\cite{simonyan2013deep,shrikumar2017learning,sundararajan2017axiomatic,smilkov2017smoothgrad,feng2018pathologies}, perturbation-based methods~\cite{li2016understanding}, local approximation methods~\cite{ribeiro2016should,ribeiro2018anchors}, prototype methods~\cite{chen2019looks,hase2019interpretable}, counterfactual and decision boundary methods~\cite{joshi2018xgems,samangouei2018explaingan, mothilal2020dice}, natural language generation~\cite{camburu_e-snli:_2018, rajani_explain_2019, hase2020leakage, wiegreffe2020}, and recently, $k$NN-based methods~\cite{papernot_deep_2018, rajani2020explaining}.

\paragraph{Influence Functions.}
The use of influence-based diagnostics can be traced back to the seminal papers such as~\citet{cook1977detection,cook1980characterizations,cook1982residuals,cook1986assessment}.
Recently,~\citet{koh2017understanding} brought influence functions to large-scale deep learning and have been followed up by numerous publications. For example,~\citet{koh2018stronger} used them for data poisoning attacks,~\citet{schulam2019can} for calibrating trust in individual model predictions,~\citet{brunet2019understanding} for tracing biases in word embeddings,~\citet{koh2019accuracy} and~\citet{basu2020second} for identifying important groups of training data, and~\citet{feldman2020neural} for studying neural networks memorization.

In the context of influence functions for NLP,~\citet{han2020explaining} used them to explain model predictions and unveil data artifacts.~\citet{yang2020g} used them to estimate the quality of synthetic training samples in the context of data-augmentation.~\citet{Meng2020PairTD} explored the combination of gradient-based methods and influence functions to jointly examine training history and test stimuli. Their work also tried using influence functions to fix erroneously classified examples, albeit through retraining another model.
In our work, we primarily focus on improving the run-time of the influence function, and explore its potential in interpreting model behaviors and efficiently correcting predictions.~\citet{kobayashi-etal-2020-efficient} experimented with training models with instance-specific dropout masks
to efficiently compute the influence of one training data-point on one test data-point. Our method, on the other hand, both speeds up individual influence function computation (Sec.~\ref{subsec:speedup-ihvp}) and reduces the number of such computations (Sec.~\ref{subsec:speedup-argmax}). Further, it
is an inference-time technique; it requires no change to model training and could, in principle, work with any trained model. Finally, we also include a more extensive set of experiments/applications on a larger dataset.

\section{\textsc{FastIF}: Method Details}

\begin{table*}
% \small
\centering
\begin{tabular}{c|c|c|c}
        & \textbf{Original $s_{\textrm{test}}$ (1 GPU)} & \textbf{Fast $s_{\textrm{test}}$ (1 GPU)}   & \textbf{Fast $s_{\textrm{test}}$ (4 GPUs)} \\
\hline
\textbf{No $k$NN} & $\geq 2 \textrm{ hours } (1\textrm{X})$ & / & / \\
\textbf{$k$NN $k=1e4$} & $14.72$ $\pm 0.21 \textrm{ min } (8\textrm{X})$ & $6.61 \pm 0.03 \textrm{ min } (18\textrm{X})$ & $2.36 \pm 0.02 \textrm{ min } (50\textrm{X})$ \\
\textbf{$k$NN $k=1e3$} & $10.93 \pm 0.04 \textrm{ min } (10\textrm{X})$ & $3.13 \pm 0.05 \textrm{ min } (38\textrm{X})$ & $1.46 \pm 0.07 \textrm{ min } (82\textrm{X})$
\end{tabular}
\vspace{-7pt}
\caption{Averages and standard deviations of influence functions runtime measured on $10$ MultiNLI examples.
}
\label{table:speed-comparisons}
\vspace{-10pt}
\end{table*}

\subsection{Background}
Let a data-point be defined as $z {=} (x, y)$ for input $x$ and label $y$, and the loss function be $L (z, \theta)$.
Given $N$ training data-points in the training set $\mathcal{Z}$, the standard empirical risk minimization tries to solve the following problem, $\hat{\theta} = \argmin _{\theta} \frac{1}{N} \sum_{i=1}^{N} L(z_i, \theta)$.
We want to answer the question: what is the influence of a training data-point $z$ on the learned model parameters $\theta$ and its behavior on a new test data-point $z_{\textrm{test}}$.

Leave-one-out methods take the ``discrete'' way: train two models, one on the full training dataset $\mathcal{Z}$, and one on the same dataset but with $z$ removed. The difference in the behavior between these two models is the effect of $z$'s presence (or absence). Among the many definitions of model behavior, in this work we mainly focus on the loss at the test data-point.
Influence functions, on the other hand, answer this question via approximating it locally and measure the change in the model behavior via up-weighting the loss of the training data-point by $\epsilon$. Thus the influence function refers to \emph{the change in the model's loss on the test data-point $z_\textrm{test}$ if we up-weight the loss of training data-point $z$ by $\epsilon$,} $\mathcal{I} (z, z_{\textrm{test}}) := \frac{dL(z_{\textrm{test}}, \hat{\theta}_{\epsilon, z})}{d\epsilon}$.
where $\hat{\theta}_{\epsilon, z}$ are the parameters of the model trained with training data-point $z$ up-weighted by $\epsilon$, $\hat{\theta}_{\epsilon, z} = \argmin _{\theta} \frac{1}{N} \sum_{i=1}^{N} \big( L(z_i, \theta) + \epsilon L(z, \theta) \big)$.

Measuring $\mathcal{I} (z, z_{\textrm{test}})$ via training another model is prohibitively expensive. The influence function~\cite{koh2017understanding} computes the following tractable approximation,
\begin{equation}
    \mathcal{I}(z, z_{\textrm{test}}) {\approx} {-}\nabla_{\theta} L (z_{\textrm{test}}, \hat{\theta})^T  H_{\hat{\theta}}^{-1}\nabla_{\theta}L(z, \hat{\theta})
\label{eq:influence-z-ztest}
\end{equation}
where $\hat{\theta}$ is the original parameter vector of model trained using training data-points, and $H_{\hat{\theta}}$ is the Hessian of model parameters. For each test data-point $z_{\textrm{test}}$, we are usually interested in finding the most positively influential training data-points, and the most negatively influential training data-points.\footnote{A training data-point is positively influential (or ``harmful'') w.r.t. a test data-point if up-weighting its loss leads to higher loss on the test data-point. A negatively influential data-point (or ``helpful'') is defined similarly.}
We can find the most positively influential (i.e., harmful) training data-points $z^*$ by computing the influence values between $z_{\textrm{test}}$ and each $z$ in the training set,
\begin{equation}
z^* = \argmax\nolimits_{z \in \mathcal{Z}} \, \mathcal{I} (z, z_{\textrm{test}})
\label{eq:argmax-influence}
\end{equation}
where $\argmax$ changes to $\argmin$ for finding the most negatively influential (i.e., helpful) data-point. While a much more tractable approximation, this is still unscalable in practice for a few reasons.

First, enumerating all the data-points in the training set to find the $\argmax$/$\argmin$ influence values is expensive for large datasets. Second, evaluating the inverse-Hessian $H^{-1}_{\hat{\theta}}$ is very expensive for large neural-network models~\cite{koh2017understanding,agarwal2017second}. 
In this work, we address those challenges by presenting simple but effective methods. We summarize the speed improvements in MultiNLI settings in Table~\ref{table:speed-comparisons} and Sec.~\ref{subsec:computation-times}, and our method in Fig.~\ref{figure:main}.

\subsection{Speeding up the $\argmax$ using $k$NN}
\label{subsec:speedup-argmax}

A naive implementation of Eq.~\ref{eq:argmax-influence} is expensive because the computation cost grows with the dataset size.
In practice, however, we are primarily concerned with data-points that are most influential. We hypothesize that we could constrain the expensive search to a subset of promising data points, $\hat{\mathcal{Z}} \subseteq \mathcal{Z}$, that are likely to be influential, without a significant impact on quality,
\begin{equation}
    z^* = \argmax\nolimits_{z \in \hat{\mathcal{Z}}} \, \mathcal{I} (z, z_{\textrm{test}})
\end{equation}
Notice that the $\arg\max$ is now performed over $\hat{\mathcal{Z}} \subseteq \mathcal{Z}$. We select the subset $\hat{\mathcal{Z}}$ as the top-$k$ nearest neighbors of $z_{\textrm{test}}$ based on the $\ell_2$ distance between extracted features of the data-points following~\citet{khandelwal2019generalization} and~\citet{rajani2020explaining}.\footnote{We use the model's final representation as the features.} This operation is extremely fast using highly-optimized nearest neighbor search libraries such as FAISS~\cite{JDH17}. 
In Sec.~\ref{subsec:knn-recalls}, we will examine the quality of nearest neighbors in terms of retrieving influence-worthy data-points.

\subsection{Speeding up the Inverse Hessian}
\label{subsec:speedup-ihvp}
The next computational bottleneck lies in estimating the inverse Hessian of model parameters in Eq.~\ref{eq:influence-z-ztest}.
Computing the Hessian for the full training dataset is expensive, and inverting it is similarly prohibitive: with $n$ training data-points and $p$ parameters, this computation requires $\mathcal{O} (np^2 + p^3)$ operations, and is very expensive for large dataset/model (please see Sec.~3 in ~\citet{koh2017understanding} for details). We start by describing the method proposed in~\citet{koh2017understanding}.

First, notice that for each test data-point $z_\textrm{test}$, we can pre-compute and cache the following quantity,
\begin{equation}
    s_{\textrm{test}} = H_{\hat{\theta}}^{-1} \nabla_{\theta} L(z_{\text {test }}, \hat{\theta})
\end{equation}
We can then efficiently compute the influence for each data-point $z$, 
$\mathcal{I}(z, z_\textrm{test}) {=} {-}s_{\textrm{test }} \cdot \nabla_{\theta} L(z_{i}, \hat{\theta})$.
Next, the method approximates $s_{\textrm{test}}$ via a combination of (1) implicit Hessian-vector products (HVPs) to avoid explicitly computing/storing $H_{\hat{\theta}}^{-1}$, (2) using a mini-batch of data-points to obtain a stochastic unbiased estimate of the Hessian, and (3) Neumann Series (Sec~3.1 in~\citet{agarwal2017second}) to iteratively estimate the inverse. We refer interested readers to~\citet{agarwal2017second} for details regarding the LiSSA (Linear time Stochastic Second-Order Algorithm) algorithm and to Sec.~4.2 in~\citet{koh2017understanding} for details regarding non-convexity and non-convergence. The algorithm can be summarized as:
\begin{itemize}[leftmargin=*, noitemsep]
\setlength\itemsep{0em}
    \item \textbf{Step 1.} Let $v := \nabla_{\theta} L (z_{\textrm{test}}, \hat{\theta})$, and initialize the inverse HVP estimation $\hat{H}_{0,\hat{\theta}}^{-1} v = v$.
    \item \textbf{Step 2.} For $j \in \{1, 2, ..., J\}$, recursively compute the inverse HVP estimation using a batch size $B$ of randomly sampled data-point(s) $z$, $    \hat{H}_{j,\hat{\theta}}^{-1} v = v + \big(I - \hat{\nabla}^2_{\theta} L (z, \hat{\theta})\big) \hat{H}_{j-1,\hat{\theta}}^{-1} v$,
    where $J$ is a sufficiently large integer so that the above quantity converges.
    \item \textbf{Step 3.} Repeat Step 1-2 $T$ times independently, and return the averaged inverse HVP estimations.
\end{itemize}

We can see that the computation cost and approximation quality depends on: $J$ the number of recursive iterations, $T$ the number of independent runs, and $B$ the batch size. Typically, $J$ is chosen such that the approximation converges, the number of repetition is simply set $T{=}1$, and the batch size is set to the largest value that the GPU can afford.
Experiments in Sec.~\ref{subsec:hessian-speed-quality-tradeoff} examine the speed-quality trade-off under various configurations.
Contrary to the common settings, the experimental observations lead us to propose a few simple changes: (1) choose a $J$ so that approximation converges; (2) choose a small batch size; (3) make up for the noisiness of small batch size using larger $T$, which can be distributed over multiple GPUs (Sec.~\ref{subsec:parallel}).
In our experiments, we pick $J {\in} \{1000, 1500, 2000\}$ based on convergence, we found that even $B{=}1$ suffices, and we choose $T {\in} [1, 4]$.

\begin{figure}
\includegraphics[width=0.9\linewidth]{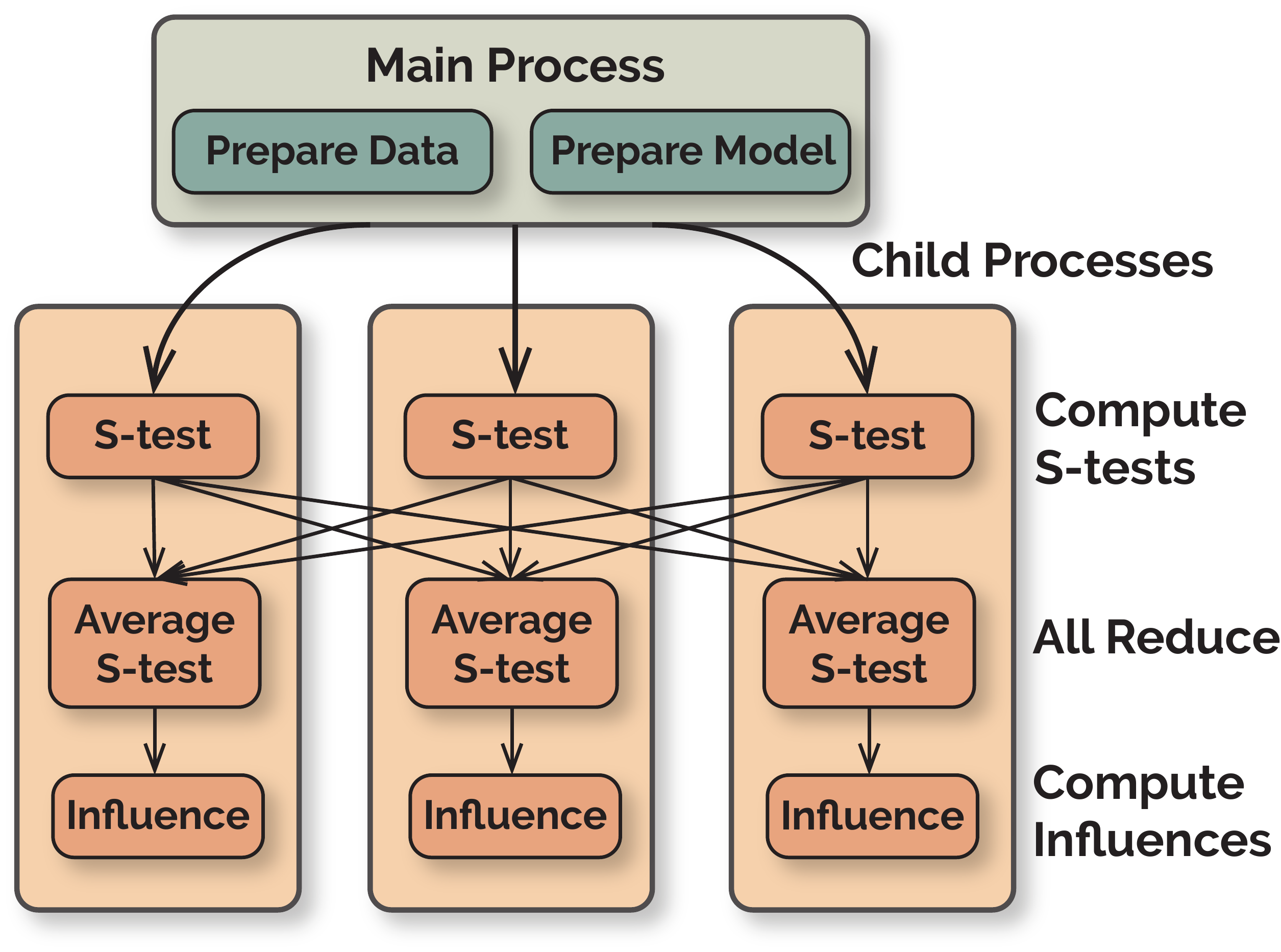}
\centering
\vspace{-11pt}
\caption{Parallel computation of influence functions.}
\vspace{-10pt}
\label{figure:parallel}
\end{figure}

\subsection{Details on Parallelization (Fig.~\ref{figure:parallel})}
\label{subsec:parallel}
The modifications described in Sec.~\ref{subsec:speedup-argmax} and~\ref{subsec:speedup-ihvp} are designed with parallelizability taken into account. Notably, the advantage of using multiple repetitions in Sec.~\ref{subsec:speedup-ihvp} is that it allows parallel-computation. We can apply asynchronous parallel computation to compute the $s_{\textrm{test}}$, use one synchronization point to average the result using \texttt{all-reduce}, and then asynchronously compute the influence of a subset of data-points that are pre-selected using $k$NN.

\section{Experimental Setup}
\label{sec:experimental-setup}
We use the MultiNLI dataset in analysis experiments (Sec.~\ref{sec:analysis}), and we include MultiNLI, HANS, ANLI, and Amazon-WILDS datasets (English) in the applications (Sec.~\ref{sec:experiments})~\cite{mccoy2019right,williams2018broad,nie2020adversarial,koh2020wilds}.\footnote{Please see corresponding papers for dataset statistics. When a MultiNLI model is intended to be used in both MultiNLI and HANS experiments, it is trained on the 2-label variant of MultiNLI. Please see Sec.~4 and Appendix~D in~\citet{mccoy2019right} for details.} Our model is based on pre-trained BERT-base model fine-tuned on downstream tasks~\cite{devlin2019bert,wolf2019huggingface}. We freeze the first 9 layers based on results in~\citet{lee2019would} to reduce computation cost, which leaves us with about $14.7$M trainable parameters. We use weight decay of $0.005$ during training as recommended by~\citet{basu2020influence}.
Please see the appendix for experiment details.
Table~\ref{table:sample-size} in the appendix summarizes key experiment details such as the number of evaluation data-points and repetitions used in experiments. We use V100 GPUs in experiments. 

\begin{figure*}
\includegraphics[width=1.0\linewidth]{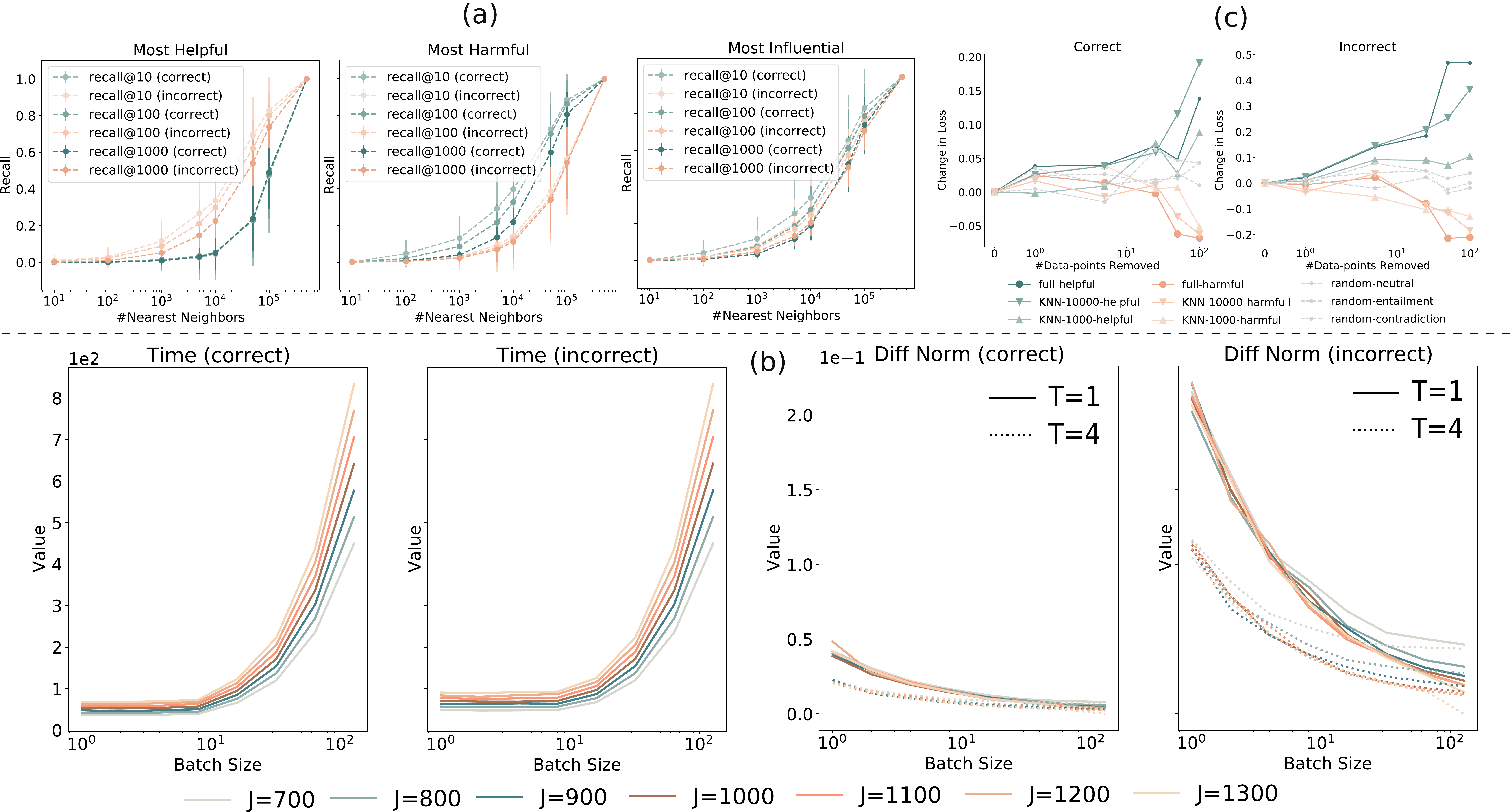}
\centering
\vspace{-15pt}
\caption{
Please see Appendix Figs.~\ref{appendix-figure:analysis-knn-recalls},\ref{appendix-figure:analysis-hessian-speed-quality-tradeoff},\ref{appendix-figure:retraining} for higher resolution versions.
\textbf{(a)} The recall of $k$NN in terms of finding influential data-points. The lines/error bars represent the means/standard deviations across $100$ correct/incorrect predictions.
\textbf{(b)} Computational time (left) and estimation error norm (right) of Hessian approximation.
\textbf{(c)} Change in loss on the data-point after retraining, where we remove $m_{\text{remove}} {\in} \{1, 5, 25, 50, 100\}$ data-points.
}
\label{figure:analysis-knn-recalls}
\label{figure:analysis-hessian-speed-quality-tradeoff}
\label{figure:retraining}
\vspace{-10pt}
\end{figure*}

\section{Experimental Results and Analysis}
\label{sec:analysis}

\subsection{Summary of Computation Times}
\label{subsec:computation-times}
Table~\ref{table:speed-comparisons} presents the summary of computation times. The run-time statistics are computed over $10$ evaluation examples on the full MultiNLI training dataset (please see Sec.~\ref{sec:experimental-setup} for details). We can see that adding $k$NN helps reduce the time by about an order of magnitude, fast $s_{\textrm{test}}$ cuts the time by additional $55{-}70\%$, and parallelism further reduces the time by more than two fold. With $k{=}1e3$, fast $s_{\textrm{test}}$ approximation, and $4$ GPUs, we are able to find influential training data-points of an evaluation data-point in about $1.5$ minutes, more than $80$ times faster than the original influence functions, which would take more than $2$ hours.

\subsection{Recall of $k$NN}
\label{subsec:knn-recalls}
To examine $k$NN's recall, we ask: \textit{if a data-point is influential, will it be included in the subset selected by the $k$NN}? We define the recall score $R@m$ as the percentage of top-$m$ ground-truth influential data-points that are selected by the $k$NN,
where we let the top-$m$ influential data-points computed without $k$NN (i.e., on the full dataset) be the top-$m$ ground-truth influential data-points.

\paragraph{Results.}
Fig.~\ref{figure:analysis-knn-recalls}~(a) shows the experiments results. Overall, the recall scores show that $k$NN is useful in selecting data-points that are likely to be influential. The recall scores for the most influential (i.e., using the absolute value) data-points are close to $60\%$ with $k{=}5 {\times} 10^4$, and $20\%$ with $k{=}5 {\times} 10^3$. These $k$'s are about order(s) of magnitude smaller than the size of the full training dataset (more than $3.9 {\times} 10^5$ examples). Note that random selection would lead to recall around $15\%$ ($k{=}5 {\times} 10^4$) and $1.5 \%$ ($k{=}5 {\times} 10^3$) in the two settings. One interesting observation is that, the recall scores for the most harmful data-points tend to be higher than the most helpful ones when the predictions are correct, but lower when the predictions are incorrect.\footnote{Experiments in Sec.~\ref{subsec:effect-visualization} find that when the predictions are incorrect, there tend to exist a few very-helpful data-points and many more relatively-medium harmful ones. This might help explain that $k$NN tends to have higher recall on helpful data-points when the predictions are incorrect.}

\subsection{Inverse HVP Approximation}
\label{subsec:hessian-speed-quality-tradeoff}
We look at the speed-quality trade-off of the Hessian approximation. Our experiments show that the speed-up does not greatly affect the quality.

\paragraph{Results.}
Fig.~\ref{figure:analysis-hessian-speed-quality-tradeoff}~(b) shows the results of the experiments. For the two figures on the left, we can observe that the computation cost (measured in time) grows with both the batch size and number of recursive iterations $J$. Similarly, the estimation error
\footnote{The estimation error is measured as the difference norm w.r.t. the estimation using the most expensive configuration.}
(the figures on the right) shows that the error decreases with both the batch size and $J$ in general.

Further, we notice that when the batch size is small (e.g., $B{=}1$), we can make up the loss in quality by increasing $T$, which is inherently parallelizable (Sec.~\ref{subsec:parallel}). Overall, these suggest that we could trade off a small drop in quality with a significant speed-up by the combination of (1) small $B$, (2) medium $J$, and (3) large $T$.

\begin{table}
\centering
\small
\begin{tabular}{l | p{0.07\textwidth}p{0.07\textwidth}p{0.07\textwidth}}
\textbf{Comparison}        & \textbf{Pearson}     & \textbf{Spearman} & \textbf{Kendall} \\
\toprule
Fast ($k{=}10^3$) vs Full            & $99.8 {\pm} 0.3$ & $99.5 {\pm} 1.3$ & $96.9 {\pm} 2.8$ \\
Fast ($k{=}10^4$) vs Full            & $99.9 {\pm} 0.1$ & $99.7 {\pm} 0.6$ & $96.8 {\pm} 2.3$ \\
Fast ($k{=}10^3$ vs $10^4$) & $99.9 {\pm} 0.2$ & $99.6 {\pm} 0.8$ & $96.6 {\pm} 2.4$ \\
\bottomrule
\end{tabular}
\vspace{-10pt}
\caption{Correlations between influence values.}
\vspace{-12pt}
\label{table:influence-correlations}
\end{table}

\subsection{Quality of Influence Estimations}
\label{subsec:retraining}
Finally, we want to ensure that the final computed influence scores are of sufficient quality.
First, we compute the correlations of influence values between one that computes influence functions with both $k$NN and fast $s_\textrm{test}$ approximation (\textbf{Fast}) and one without them (\textbf{Full}). 
Next, we compare the quality of computed influences by actually retraining the models.
If the influence function correctly identifies helpful and harmful data-points, retraining without helpful points should result in the evaluation loss rising (i.e., positive change in loss), and retraining without harmful points should result in the loss falling (i.e., negative change in loss).

\paragraph{Results.}
Table~\ref{table:influence-correlations} shows that the correlations are high for all measures considered. The results show that fast influence functions achieve ${>}95\%$ correlations with full influence functions. These demonstrate that using the fast influence functions achieve reasonable qualities at just a fraction of the total cost.
Next, Fig.~\ref{figure:retraining}~(c) shows the retraining results, where we separate the results for cases based on whether the prediction is correct, and averaged across data-points within this bucket.
We can see that overall loss increases when we remove helpful data, decreases when we remove harmful data-points. Further, the performance (measured by the change in loss) between using the fast influence functions and full influence functions (i.e., no $k$NN and fast $s_\textrm{test}$ approximation) is similar, and both perform better than random selection in general. Note that with a large training set, individual data points tend to have a limited effect on the generalization error. We see a larger effect as a larger number of points are removed.

\section{Applications of \textsc{FastIF}}
\label{sec:experiments}

\begin{figure*}
\includegraphics[width=0.99\textwidth]{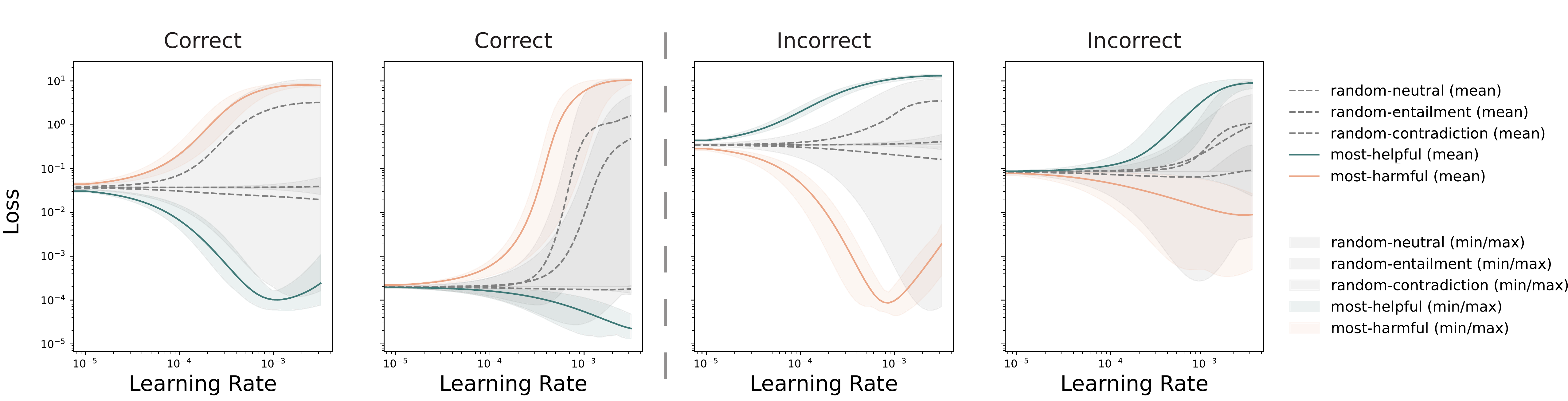}
\centering
\vspace{-9pt}
\caption{Simulator loss on $4$ test data-points (more figures in the appendix), where the simulator is fine-tuned on different types of data-points with ground truth labels using various learning rates.
The lines refer to the mean performance averaged across $10$ fine-tuning data-points, and the shaded area covers the max/min performance.}
\label{figure:task-model-simulator-model}
\vspace{-5pt}
\end{figure*}

\subsection{Explainability of Influential Examples}
\label{subsec:explainability-of-influential-examples}
\paragraph{Motivation.}
Knowing which points are influential to the model loss may give us insight into our model/training algorithm, and therefore we want to test whether influence functions can be used to improve model explainability. To do so, we need a framework for evaluating the quality of explanations given in terms of influential data points.

\paragraph{Approach.}
Here we will mainly focus on the concept of \textit{simulatability}.~\citet{doshi2017towards} explained that a model is simulatable when a person can predict its behavior on new inputs. Thus, we will measure an explanation's quality in terms of its effect on model simulatability.
Specifically, we train another \textit{simulator} model to \textit{predict the predictions} of the \textit{task} model~\cite{treviso-martins-2020-explanation, hase2020leakage}. The simulator model is trained with the same data as the task model, but the labels are replaced with the task model's predictions. We then fine-tune the simulator on data identified as influential for the task model's performance on test data-points. If the simulator model can better predict what the task model will do by fine-tuning on this data (i.e., achieving lower loss), the influential data points are said to be a good ``explanation" of the task model's behavior. This approach is similar to~\citet{pruthi2020}, who also treat explanations as targets to fit a model to. 

\paragraph{Experiment Results.}
We can observe from Fig.~\ref{figure:task-model-simulator-model} that, when the prediction is correct, finetuning on helpful data-points improves the simulator's ability to predict the task model's behavior. 
Similarly, when the prediction is incorrect, finetuning on data-points that are \emph{harmful} (to the task model's loss on ground-truth labels) \emph{improves} the simulator's predictions of the task model.
Further, the effect on the loss from influential data-points is greater than random selection of data-points overall. This demonstrates that influential examples can serve as explanations of the task model's behavior.

\begin{figure*}[t]
\includegraphics[width=0.99\textwidth]{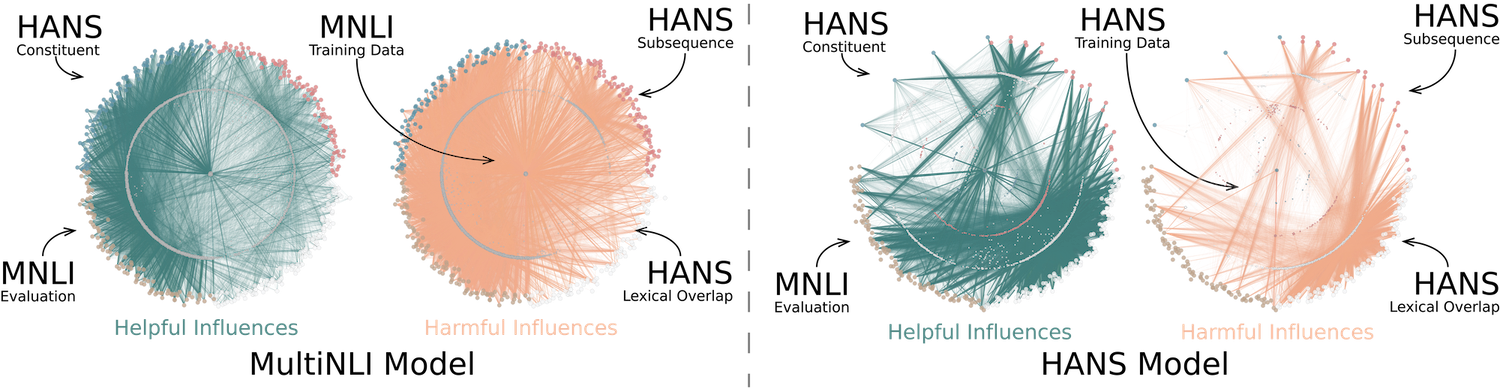}
\centering
\vspace{-9pt}
\caption{Visualization of the interaction between evaluation and training data-points.}
\vspace{-10pt}
\label{figure:effect-visualization}
\end{figure*}

\subsection{Effect Visualization}
\label{subsec:effect-visualization}
\paragraph{Motivation.}
Investigating how different training data-points interact with test data-points is also useful, because such exploratory data analysis can discover interesting relationships between data-slices (i.e., subsets of data with specific attributes).

\paragraph{Approach.}
We conduct experiments on two models, one trained on MultiNLI and the other on HANS. We then compute influence functions on their corresponding training data-points and build circular bipartite graphs. The nodes represent training (inner circle) and evaluation (outer circles) data-points (incorrect predictions), and the strength and color of edges represent the influence values.

\paragraph{Experiment Results}
Fig.~\ref{figure:effect-visualization} visualizes the results, where the left half corresponds to the model trained on MultiNLI and the right half corresponds to the model trained on HANS.

\textbf{Left Hand Side.}
Appendix Table~\ref{table:effect-visualization-table-1} summarizes a few key statistics about the plots on the left hand side.
Interestingly, we observe that while there are more MultiNLI training data-points that are harmful to HANS/MultiNLI evaluation data-points, their (harmful) influences are in general much lower than the helpful counterparts. This suggests that a few critical helpful data-points can potentially improve the model's performance on HANS. In contrast, a few harmful data-points will have a relatively lower impact on the model's performance. In Sec.~\ref{subsec:improving-model-prediction} , we will leverage this insight to improve model performance on HANS. This could also be connected to the observation we have in Sec.~\ref{subsec:knn-recalls} where the recall scores of $k$NN tend to be higher for helpful data-points for incorrect predictions.

Further, we measure the influence correlation between different data slices.
Table~\ref{table:effect-visualization-table-2} in the appendix suggests that, for the datasets considered here, if training data-points are influential (either harmful or helpful) to one of the data slices, these data-points will likely be similarly influential to other data slices (if influential at all).

\textbf{Right Hand Side.}
Since here the training data is HANS, we further segment the HANS training data-points based on the subset they are in, using different colors and radii. Interestingly, we find that training data-points in the ``Lexical Overlap'' are noticeably more influential (either helpful and harmful) to all the HANS evaluation dataset subsets. Note that, by the construction of the dataset, the ``Lexical Overlap'' heuristic includes other heuristics as special cases.
Hence, we conclude that visualization of data influences can be used to discover latent structure in datasets.

\begin{figure*}
\includegraphics[width=0.99\linewidth]{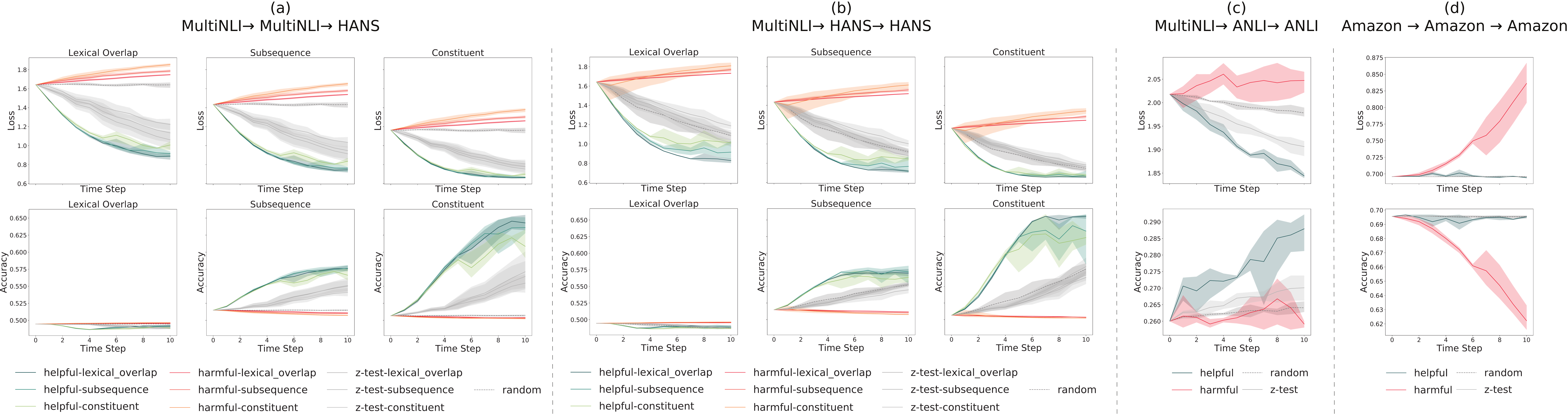}
\centering
\vspace{-3pt}
\caption{
Performance after repeatedly taking gradient steps on different types of data-points.
We use figure titles $A {\rightarrow} B {\rightarrow} C$ to refer to the settings where the model is trained on $A$, fine-tuned on $B$, and then evaluated on $C$. The lines refer to the mean performance averaged across $3$ repetitions, and the shaded area covers the max/min performance.
When models are evaluated on HANS, the influence values are computed w.r.t. each subset of HANS validation dataset (shown with different line colors), and we show results on each subset of HANS test dataset.
Please see the appendix for corresponding results on validation datasets.}
\vspace{-10pt}
\label{figure:data-augmentation-ab}
\end{figure*}

\begin{figure}
\includegraphics[width=0.99\linewidth]{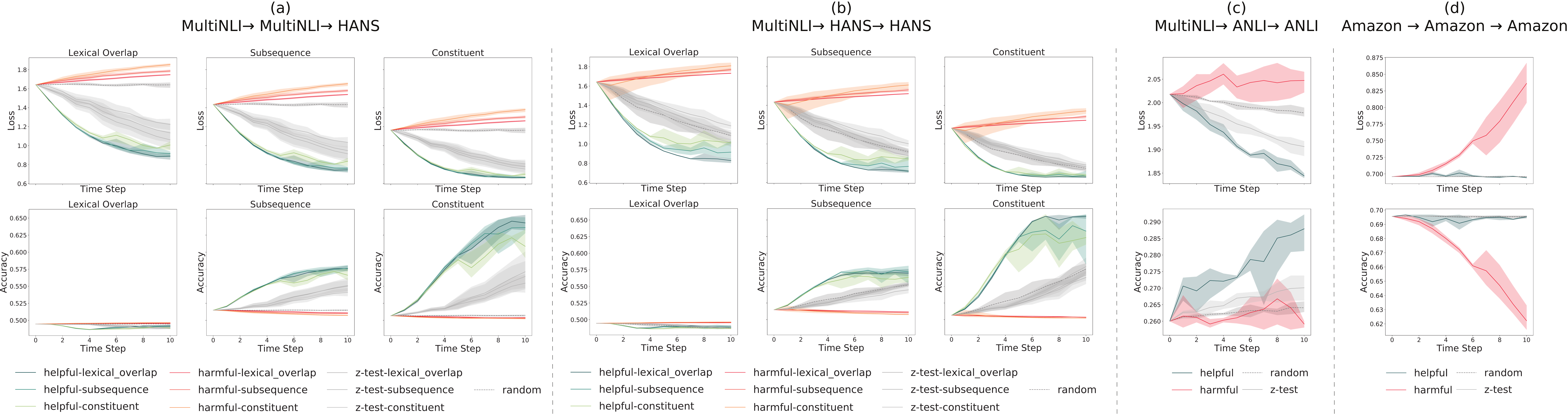}
\centering
\vspace{-20pt}
\caption{Experiments on ANLI and Amazon-WILDS.}
\vspace{-15pt}
\label{figure:data-augmentation-cd}
\end{figure}

\subsection{Error Correction}
\label{subsec:improving-model-prediction}
\paragraph{Motivation.} In addition to model interpretation and exploratory data analysis, one might also consider ways to efficiently correct the wrong model predictions, which is often useful in practice.
Luckily, the influence function not only implies which data-points are harmful,
but also which data-points that are helpful.
This suggests a simple way to correct/fix model prediction(s): taking gradient steps with the model on the helpful data-points. 
This approach naturally leads to another interesting question. Instead of taking gradient steps on data from the original data-points, can the same approach be used to suggest \textit{new} training data-points that were not seen during training? This is interesting as the original formulations of influence functions are defined on the data-points used during training.
Our experiments show that, when the new training data-points are ``similar'' to the original training data-points, this approach can be helpful. 

\paragraph{Approach.}
We first sample a small batch of validation data-points (``anchor'' data-points), compute influence scores of training data-points w.r.t. the anchor data-points, and then update model parameters by taking gradient steps on influential training data-points w.r.t. these anchor data-points. These steps are repeatedly applied multiple times. Please see Sec.~\ref{appendix-subsec:improving-model-prediction} for more details such as a step-by-step algorithm and dataset splits.

\paragraph{Experiment Results.}
Fig.~\ref{figure:data-augmentation-ab}~(a) shows the results for the model trained on MultiNLI, evaluated on HANS, and the augmented data come from the original training dataset (MultiNLI).\footnote{Note that while the losses on the lexical-overlap subset of HANS improve, the accuracies do not. This is because the predicted probabilities become more calibrated, but the output of the $\argmax$ operations do not. This still makes sense as the influence function is computed w.r.t. the loss.}
Overall, we can see that fine-tuning on helpful data-points improves the performances compared to random ones, and fine-tuning on harmful data points leads to worse performances. This simple technique brings more than $5.8\%$ average improvement in accuracy.

As using influence functions requires gradient access to the anchor data-points, we also experiment with directly fine-tuning on them as well (``z-test'' in the figure). The results demonstrate the potential of using influence functions to correct model predictions, above and beyond the improvements available from just finetuning on the anchor data-points (by about $2.5\%$ in accuracy on average).

We can also observe from Fig.~\ref{figure:data-augmentation-ab}~(a) that helpful examples tend to have a greater magnitude of effect than harmful examples. This can be connected to the visualization results in Sec.~\ref{subsec:effect-visualization} where we can see (a handful of) helpful data-points have large negative/helpful influences while many data-points have medium positive/harmful influences.

Next, we examine the settings where we fine-tune the model on a new training dataset unseen during training instead of on the original training dataset (i.e., a data augmentation setup). Figs.~\ref{figure:data-augmentation-ab}~(b) and~\ref{figure:data-augmentation-cd}~(c) show the results for the model trained on MultiNLI, evaluated on HANS/ANLI, and the augmented data come from the HANS/ANLI training dataset.
We can observe that random data augmentation works reasonably well.\footnote{This intuitively makes sense because the evaluation data comes from HANS/ANLI evaluation dataset.} Further, augmenting helpful data-points can outperform random data augmentation and using anchor data-points directly in general.
In the end, we could improve the average accuracy on HANS and ANLI by more than $5.9\%$/$2.7\%$ (about $2.8\%$/$1.7\%$ more than using anchor data-points directly) respectively.
These results show the potential of influence functions for sample-efficient data augmentation.

Finally, we experiment with settings where the model is trained/fine-tuned on the Amazon-WILDS training dataset, and evaluated on an out-of-distribution (OOD) test set. Fig.~\ref{figure:data-augmentation-cd}~(d) shows that fine-tuning on harmful data-points deteriorates the performance as expected, though fine-tuning on helpful ones has little impact on the performance. We hypothesize that the selected anchor data-points are not representative enough of the evaluation dataset, as fine-tuning on anchor data-points directly also has little impact. This shows that the usefulness of our method likely depends on the quality of chosen anchor data-points.

\section{Conclusions}
We present \textsc{FastIF} which, via simple modifications, significantly speeds up the computation to influence functions without significantly impacting their performance. Our improvements include using $k$NN to pre-select a subset of influence-worthy data-points, identifying a set of hyperparameters for the inverse HVP estimation algorithm, and a parallel implementation that minimizes communication overhead. We empirically examine the effectiveness of these modifications. Then, with the availability of fast influence functions, we demonstrate a few interesting applications that were previously intractable: (1) examining the ``explainability'' of influential data-points, (2) visualizing data influence-interactions, (3) correcting model predictions using original training data, (4) and correcting model predictions using data from a new dataset.

\section*{Acknowledgments}
We thank the reviewers and Shi Feng for helpful discussions. HG interned at Salesforce Research; PH and MB were supported by a Salesforce Research Faculty Grant, NSF-CAREER Award 1846185, DARPA YFA17-D17AP00022, and a Royster Society PhD Fellowship.

\section{Ethical Considerations}
This work presents scalable influence functions for efficient model interpretation and debugging, which would be especially useful for improving model performance for particular categories of model failures after they are identified.

Recently,~\citet{carlini2020extracting} noticed that an adversary could extract training data from large-scale language models and recover potentially sensitive information. If properly used, the tool could be helpful for checking whether a model might be vulnerable to such attacks (hence, our tool should be used to encourage the detection of such memorization-based models as opposed to being misused to exploit such models).
Finally, while the fast influence functions are more compute-efficient than alternatives like retraining and full-scale influence functions, they are nevertheless expensive operations. Thus, applying \textsc{FastIF} to large-scale datasets and models might be restricted to those who have access to adequate computing power (but in fact, this is why the main purpose of this paper is to make influence functions faster and more compute efficient).

\bibliography{citations}
\bibliographystyle{acl_natbib}

\section*{Appendix}
\appendix
\begin{table*}
\centering
\small
\begin{tabular}{l|p{0.85\linewidth}}
\toprule
Section & Details \\
\midrule
Sec.~\ref{subsec:knn-recalls} / \ref{appendix-subsec:knn-recalls} &
$100$ evaluation data-points, $k$NN with $k {\in} \{1 {\times} 10^1, 1 {\times} 10^2, 1 {\times} 10^3, 1 {\times} 10^4, 5 {\times} 10^4, 1 {\times} 10^5\}$, $R@m$ with $m {\in} \{10^1, 10^2, 10^3\}$.
\\
Sec.~\ref{subsec:hessian-speed-quality-tradeoff} / \ref{appendix-subsec:hessian-speed-quality-tradeoff} &
$8$ evaluation data-points, $J {\in} \{700, 800, ..., 1300\}$, $B {\in} \{2^0, 2^1, ..., 2^7\}$, $T {\in} \{1, 4\}$.
\\
Sec.~\ref{subsec:retraining} / \ref{appendix-subsec:retraining} &
$20$ evaluation data-points for estimating correlations, and $10$ for retraining experiments, $k$NN with $k {\in} \{10^3,10^4\}$, $m_{\text{remove}} {\in} \{1, 5, 25, 50, 100\}$.
\\
Sec.~\ref{subsec:explainability-of-influential-examples} / \ref{appendix-subsec:explainability-of-influential-examples} &
$20$ evaluation data-points, $1$ fine-tuning data-point, $50$ learning rates in log-space from $10^{-5}$ to $10^{-2.5}$, and repeat for $10$ different fine-tuning data-points.
\\
Sec.~\ref{subsec:effect-visualization} / \ref{appendix-subsec:effect-visualization} & $400$ evaluation data-points where the model predictions are incorrect, including $100$ from the MNLI evaluation dataset and $100$ for each of the three subsets from the HANS evaluation dataset. We use $k$NN with $k=10^3$.
\\
Sec.~\ref{subsec:improving-model-prediction} / \ref{appendix-subsec:improving-model-prediction} &
Experiments are repeated $3$ times, fine-tuning is applied repeatedly for $10$ iterations. For each iteration, we select $10$ validation data-points as the ``anchor'' data-points, update parameters for one gradient step on $10$ fine-tuning data-points with learning rate $10^{-4}$. For Amazon-WILDS experiments, we use $50$ anchor and fine-tuning data-points.
\\
\bottomrule
\end{tabular}
\vspace{-10pt}
\caption{Some of the key experiment details.
}
\vspace{-10pt}
\label{table:sample-size}
\end{table*}

\section{Summary of Key Experiment Details}
Please see Table~\ref{table:sample-size}.

\section{Experimental Results and Analysis}
\label{appendix-sec:analysis}

In this section, we examine the effectiveness of the methods described in Sec.~\ref{subsec:speedup-argmax} and~\ref{subsec:speedup-ihvp}. Specifically, we conduct the following set of experiments:
\begin{itemize}[leftmargin=*, noitemsep]
    \item We measure $k$NN's recall in terms of retrieving data-points that are potentially influential.
    \item We look at the trade-off in speed/quality of inverse Hessian-vector-product approximation with various configurations.
    \item We examine the quality of the influence estimations using all of the proposed techniques, by comparing the correlations between the influence values computed with/without them.
\end{itemize}

\subsection{Recall of $k$NN}
\label{appendix-subsec:knn-recalls}
In Sec.~\ref{subsec:speedup-argmax}, we describe the use of $k$NN for pre-selecting a subset of data-points. This (smaller) set of data-points are then re-ranked using the more expensive influence functions. Making this work well requires that the subset selected by $k$NN contains potentially the most influential data-points. 

In this section, we examine the $k$NN's recall. We ask the question: \textit{if a data-point is influential, will it be included in the subset selected by the $k$NN}? To formalize this, we define the recall score $R@m$ as the percentage of top-$m$ ground-truth influential data-points that are selected by the $k$NN,
\begin{equation*}
    R@m = \frac{\mid\{\text { retrieved }\} \cap\{\text { top-$m$ influential }\} \mid}{\mid\{\text { top-$m$ influential }\} \mid}
\end{equation*}
where we let the top-$m$ influential data-points computed without $k$NN (i.e., on the full dataset) be the top-$m$ ground-truth influential data-points.

\paragraph{Details.}
For each evaluation data-point $z_\textrm{test}$, we first compute the ground-truth influential data-points via running influence functions on the MultiNLI training dataset without $k$NN (i.e., $\{\text { top-$m$ influential }\}$). Then, we use $k$NN to select the $k$ training data-points (i.e., $\{\text { retrieved }\}$). We chose $k {\in}$ $\{1 {\times} 10^1, 1 {\times} 10^2, 1 {\times} 10^3, 1 {\times} 10^4, 5 {\times} 10^4, 1 {\times} 10^5\}$. Finally, we compute the recall $R@m$ with $m {\in} \{10^1, 10^2, 10^3\}$.
We repeat the aforementioned steps for three types of influential data-points: most positively influential (harmful), most negatively influential (helpful), and most influential (unsigned influence, by taking the absolute value). We select $100$ data-points from the MNLI evaluation dataset ($50$ data-points when the model predictions are correct and incorrect) and aggregate the results.

\begin{figure*}
\includegraphics[width=0.99\linewidth]{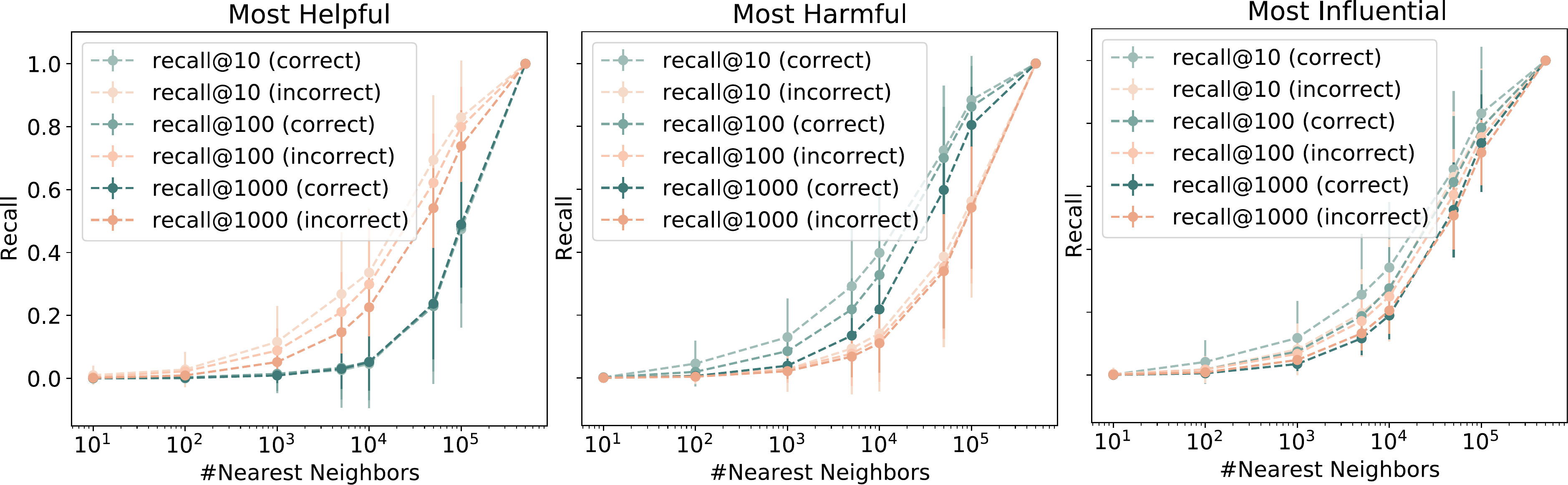}
\centering
% \vspace{-15pt}
\caption{The recall of $k$NN in terms of finding influential data-points. The lines and error bars represent the means and standard deviations across $100$ correct/incorrect predictions.
}
\label{appendix-figure:analysis-knn-recalls}
% \vspace{-15pt}
\end{figure*}

\paragraph{Results.}
Fig.~\ref{appendix-figure:analysis-knn-recalls} shows the experiments results. Overall, the recall scores show that $k$NN is useful in selecting data-points that are likely to be influential. The recall scores for the most influential (i.e., using the absolute value) data-points are close to $60\%$ with $k{=}5 {\times} 10^4$, and $20\%$ with $k{=}5 {\times} 10^3$. These $k$'s are about order(s) of magnitude smaller than the size of the full training dataset (more than $3.9 {\times} 10^5$ examples). Note that random selection would lead to recall around $15\%$ ($k{=}5 {\times} 10^4$) and $1.5 \%$ ($k{=}5 {\times} 10^3$) in the two settings. One interesting observation is that, the recall scores for the most harmful data-points tend to be higher than the most helpful ones when the predictions are correct, but lower when the predictions are incorrect.\footnote{
Experiments in Main Paper Sec.~\ref{subsec:effect-visualization} find that when the predictions are incorrect, there tend to exist a few very-helpful data-points and many more relatively-medium harmful ones. This might help explain that $k$NN tends to have higher recall on helpful data-points when the predictions are incorrect.}

\subsection{Inverse-Hessian-Vector-Product Approximation Speed-Quality Trade-Off}
\label{appendix-subsec:hessian-speed-quality-tradeoff}
We look at the speed-quality trade-off of the Hessian approximation tricks, and see if the Hessian approximation's speed-up comes at the cost of dramatically lower quality. Our experiments show that the speed-up does not greatly affect the quality.

\paragraph{Details.}
We compute the Hessian approximations with $J {\in} \{700, 800, ..., 1300\}$, $B {\in} \{2^0, 2^1, ..., 2^7\}$, $T {\in} \{1, 4\}$,\footnote{In the figures, we only include the results of different $T$ for difference norm. This is because practitioners can use parallelism to speed up the computations of each run, which is independent of each other as described in Sec.~\ref{subsec:parallel}. Thus, the change in time mainly depends on parallelism overhead, which we found to be reasonable in our initial experiments.} and repeat the experiments for $8$ different MultiNLI evaluation data-points ($4$ for correct/incorrect predictions).

\begin{figure*}
\includegraphics[width=0.95\linewidth]{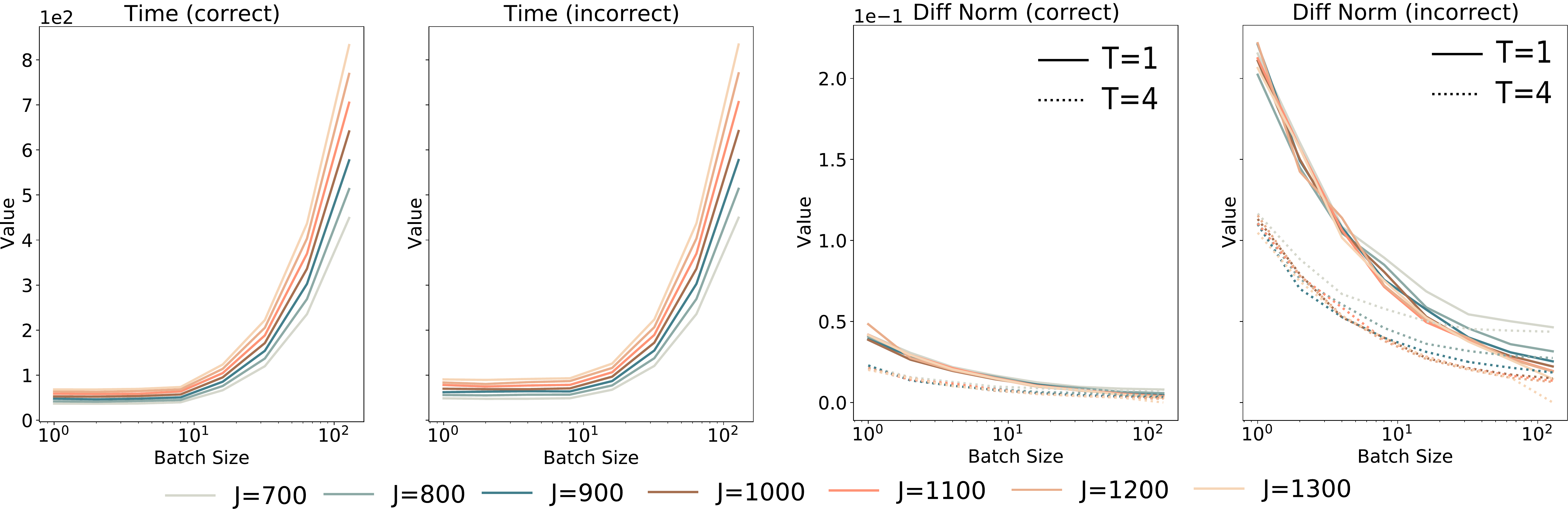}
\centering
% \vspace{-7pt}
\caption{
\textbf{Left Half:} computational time of Hessian approximation as a function of batch size and recursive iterations $J$. We further break down the figure into two sub-figures: cases when the prediction is correct and those when it is incorrect. \textbf{Right Half:} estimation error norm as a function of batch size, recursive iterations, whether the prediction is correct, and additionally the number of independent runs $T$.
}
% \vspace{-7pt}
\label{appendix-figure:analysis-hessian-speed-quality-tradeoff}
\end{figure*}

\paragraph{Results.}
Fig.~\ref{appendix-figure:analysis-hessian-speed-quality-tradeoff} shows the results of the experiments. For the two figures on the left, we can observe that the computation cost (measured in time) grows with both the batch size and number of recursive iterations $J$. Similarly, the estimation error\footnote{The estimation error is measured as the difference norm w.r.t. the estimation using the most expensive configuration.} (the figures on the right) shows that the error decreases with both the batch size and $J$ in general.

Further, we notice that when the batch size is small (e.g., $B{=}1$), we can make up the loss in quality by increasing $T$, which is inherently parallelizable (Main Paper Sec.~\ref{subsec:parallel}). Overall, these suggest that we could trade off a small drop in quality with a significant speed-up by the combination of (1) small $B$, (2) medium $J$, and (3) large $T$.\footnote{We pick $J {\in} \{1000, 1500, 2000\}$ based on convergence.}

\begin{table}
\centering
\scriptsize
\begin{tabular}{l | lll}
\textbf{Comparison}        & \textbf{Pearson}     & \textbf{Spearman} & \textbf{Kendall} \\
\toprule
Fast ($k{=}10^3$) vs Full            & $99.8 {\pm} 0.33$ & $99.5 {\pm} 1.34$ & $96.9 {\pm} 2.81$ \\
Fast ($k{=}10^4$) vs Full            & $99.9 {\pm} 0.10$ & $99.7 {\pm} 0.55$ & $96.8 {\pm} 2.27$ \\
Fast ($k{=}10^3$ vs $k{=}10^4$) & $99.9 {\pm} 0.18$ & $99.6 {\pm} 0.75$ & $96.6 {\pm} 2.35$ \\
\bottomrule
\end{tabular}
% \vspace{-7pt}
\caption{Correlations between influence values using various measures. The means and standard deviations are computed with $20$ evaluation points 
(balanced between correct and incorrect predictions).
}
% \vspace{-15pt}
\label{appendix-table:influence-correlations}
\end{table}

\subsection{Quality of Influence Estimations}
\label{appendix-subsec:retraining}
Finally, we want to ensure that there are no significant cascading estimation errors and that the final computed influence scores are of sufficient quality. Think of Sec.~\ref{appendix-subsec:knn-recalls} and~\ref{appendix-subsec:hessian-speed-quality-tradeoff} as ``unit-tests'' to understand whether each component works well on its own, and in this section, our goal is to understand whether all the components work well together.

First, we compute the correlations of influence values among the following two systems: one that computes influence functions with both $k$NN and fast $s_\textrm{test}$ approximation (\textbf{Fast}, with $k{=}10^3/k{=}10^4$), and one that computes influence functions without them (\textbf{Full}). 

Next, we further compare the quality of computed influences by actually retraining the models using three systems: (1) a system that computes influence functions with both $k$NN and fast $s_\textrm{test}$ approximation, (2) a system that computes influence functions without them, and (3) a system that randomly selects data-points. 

For each evaluation data-point selected, we find its $m_{\text{remove}}$ most influential training data-points. Then we retrain the model by removing them and measure the change in the loss on the same evaluation data-point.
If the influence function correctly identifies helpful and harmful data-points, retraining without helpful points should result in the evaluation loss rising (i.e., positive change in loss), and retraining without harmful points should result in the loss falling (i.e., negative change in loss).

\begin{figure}
\includegraphics[width=0.99\linewidth]{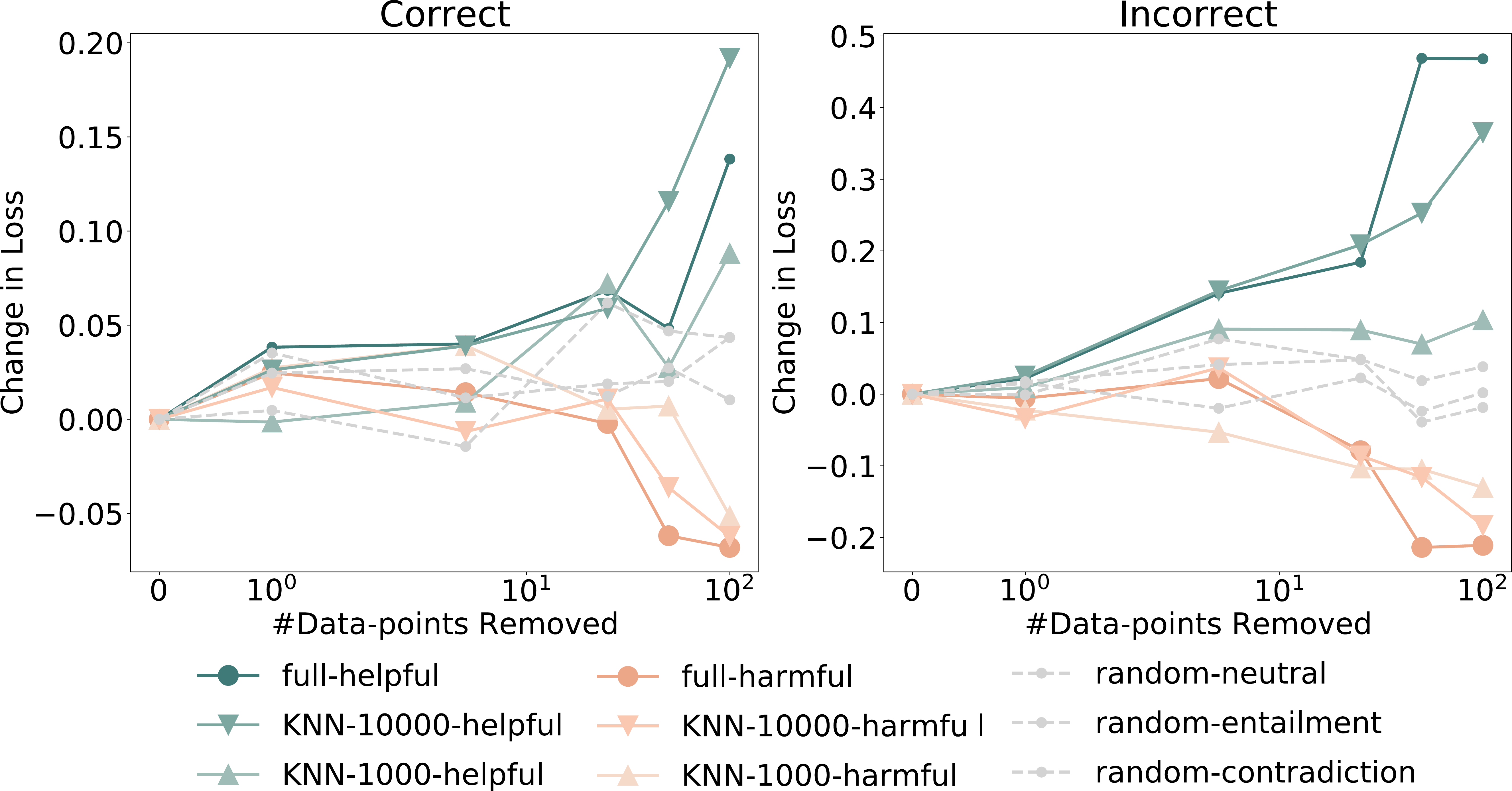}
\centering
% \vspace{-5pt}
\caption{
Change in loss on the data-point after retraining, where we remove $m_{\text{remove}} {\in} \{1, 5, 25, 50, 100\}$ data-points. We can see that the fast influence algorithms produce reasonable quality estimations at just a fraction of computation cost.
}
% \vspace{-15pt}
\label{appendix-figure:retraining}
\end{figure}

\paragraph{Details.}
For each evaluation data-point selected, we find its influential training data-points using one of the three aforementioned systems. Then we retrain the model by removing $m_{\text{remove}}$ training data-point(s) and measure the change in the loss on the same evaluation data-point.
For system (1), we use $k$NN (with $k {\in} \{10^3, 10^4$\}) and fast $s_\textrm{test}$ approximation, and for system (3), we randomly select data-points with each of the three labels.
We choose $m_{\text{remove}} {\in} \{1, 5, 25, 50, 100\}$ and repeat the experiment for $10$ data-points ($5$ data-points where the prediction is correct and incorrect) that have at least $100$ harmful/helpful data-points.

\paragraph{Results.}
First, Table~\ref{appendix-table:influence-correlations} shows that the correlations are high for all measures considered. Notably, the results show that fast influence functions achieve ${>}95\%$ correlations with full influence functions. These demonstrate that using the fast influence functions achieve reasonable qualities at just a fraction of the total cost.
Next, Fig.~\ref{appendix-figure:retraining} shows the retraining results, where we separate the results for cases based on whether the prediction is correct, and averaged across data-points within this bucket.
We can see that overall loss increases when we remove helpful data, decreases when we remove harmful data-points. Further, the performance (measured by the change in loss) between using the fast influence functions and full influence functions (i.e., no $k$NN and fast $s_\textrm{test}$ approximation) is similar, and both perform better than random selection in general.

% % 

\begin{figure*}
\includegraphics[width=0.99\textwidth]{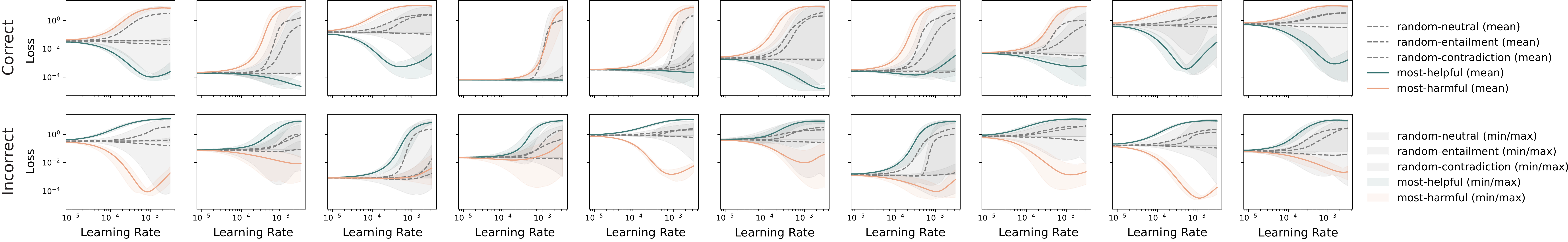}
\centering
\vspace{-5pt}
\caption{Simulator loss on $20$ evaluation data-points.}
\vspace{-10pt}
\label{appendix-figure:task-model-simulator-model-extended}
\end{figure*}

\begin{table}
\centering
\small
\begin{tabular}{l | ll | ll}
\toprule
\multirow{2}{*}{\textbf{Slice}} & \multicolumn{2}{l|}{\textbf{Average $|\mathcal{I}|$} ($\times 10^{-1}$)} & \multicolumn{2}{l}{\textbf{\# Edges ($\times 10^5$})} \\
             & Helpful & Harmful & Helpful & Harmful \\
\midrule
H (L)     & $8.66$  & $0.34$ & $0.05$ & $0.95$ \\
H (S)     & $9.34$ & $0.66$ & $0.08$ & $0.92$ \\
H (C)     & $4.65$ & $1.79$ & $0.29$ & $0.71$ \\
M         & $136.89$ & $44.19$ & $0.27$ & $0.73$  \\
\bottomrule
\end{tabular}
% \vspace{-5pt}
\caption{Statistics on the visualizations.
Harmful edges have positive influences, helpful edges have negative influences, but here we use average absolute values (\textbf{Average $|\mathcal{I}|$}).
\textbf{\#~Edges} refers to the number of (harmful/helpful) edges connecting to the slice.
``H (L/S/C)'' refers to the three subset of HANS,
and ``M'' to the 2-label version of MultiNLI. Note that for average $|\mathcal{I}|$, comparisons are only meaningful within each row.}
% \vspace{-15pt}
\label{table:effect-visualization-table-1}
\end{table}

\begin{table}
\centering
\small
\begin{tabular}{l|llll}
\toprule
\textbf{Slice} & HANS (L) & HANS (S) & HANS (C) & MNLI \\
\midrule
\multicolumn{5}{c}{\textbf{Correlation (Rounded)}} \\
\midrule
H (L)     & $1.00$  & $0.99$ & $0.99$ & $0.92$ \\
H (S)     & $0.99$ & $1.00$ & $0.99$ & $0.91$ \\
H (C)     & $0.99$ & $0.99$ & $1.00$ & $0.67$ \\
M         & $0.92$ & $0.91$ & $0.67$ & $1.00$  \\
\midrule
\multicolumn{5}{c}{\textbf{Number of Pairs}} \\
\midrule
H (L)     & $29581$  & $14917$ & $3425$ & $5380$ \\
H (S)     & $14917$ & $25145$ & $6218$ & $6270$ \\
H (C)     & $3425$ & $6218$ & $26621$ & $13972$ \\
M         & $5380$ & $6270$ & $13972$ & $72324$  \\
\bottomrule
\end{tabular}
% \vspace{-7pt}
\caption{Statistics on the correlations and number of pairs (where a training data-point have influences on both data slices) between data slices.
}
% \vspace{-17pt}
\label{table:effect-visualization-table-2}
\end{table}

\begin{figure*}
\includegraphics[width=0.99\linewidth]{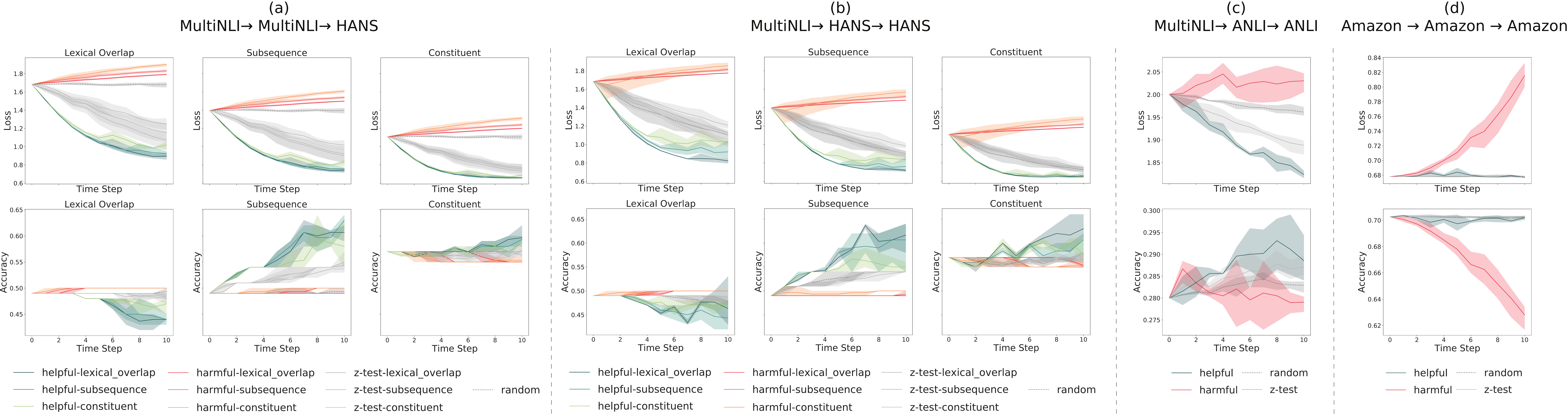}
\centering
\vspace{-7pt}
\caption{Corresponding performance on validation datasets.}
\vspace{-10pt}
\label{appendix-figure:data-augmentation-eval-ab}
\end{figure*}

\begin{figure}
\includegraphics[width=0.9\linewidth]{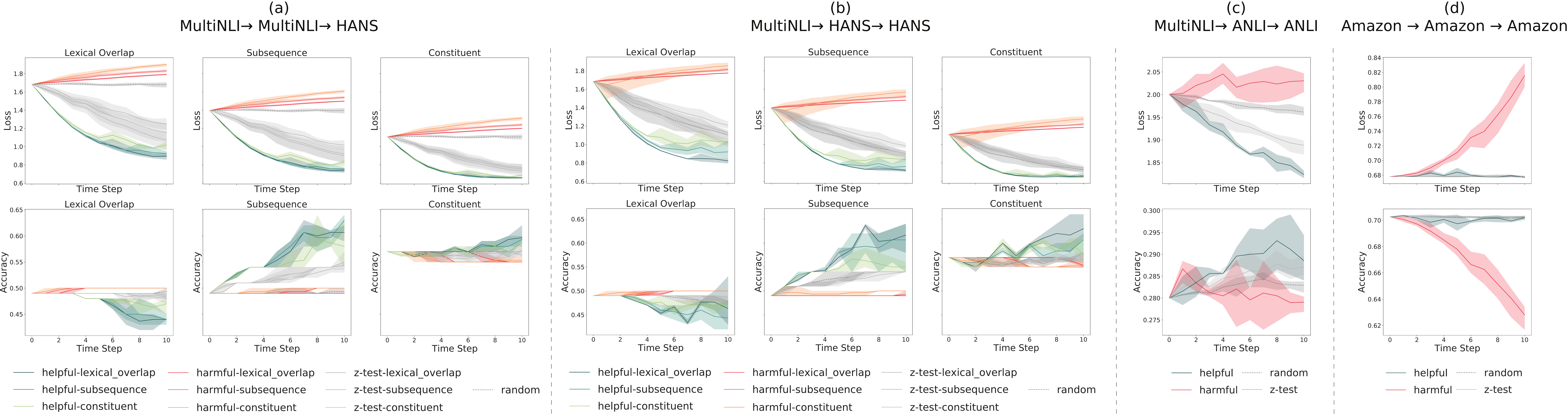}
\centering
\vspace{-7pt}
\caption{Experiments on ANLI and Amazon-WILDS.}
\vspace{-10pt}
\label{appendix-figure:data-augmentation-eval-cd}
\end{figure}

%%%%%%%%%%%%%%%%%%%%%%%%%%%%%%%%%%%%%%%%%%%%%%%%%%%%%%%%%%%%%%%%%%
% Experiments
%%%%%%%%%%%%%%%%%%%%%%%%%%%%%%%%%%%%%%%%%%%%%%%%%%%%%%%%%%%%%%%%%%

\section{Applications of \textsc{FastIF}}

\subsection{Explainability of Influential Examples}
\label{appendix-subsec:explainability-of-influential-examples}

\paragraph{Experiment Design.}
\begin{enumerate}[leftmargin=*, noitemsep]
    \item Train the ``task model'' $f_{\theta_{\textrm{task}}}$ on the original training dataset $\{x_i, y_i\}^n_{i=1}$, and another ``simulator model`` $f_{\theta_{\textrm{simulator}}}$ on the modified training dataset $\{x_i, \hat{y}_i\}^n_{i=1}$, where $\hat{y}_i {=} f_{\theta_{\textrm{task}}}(x_i)$ is the task model's prediction.
    \item Given a new test data-point $x'$ and the task model's prediction, compute the simulator model loss $\ell = L\big(f_{\theta_{\textrm{simulator}}}(x'), f_{\theta_{\textrm{task}}}(x') \big)$.
    \item Find the most influential data-point(s), and further fine-tune (i.e., take gradient steps with) the simulator model on these data-point(s), to get new parameters $\theta'_{\textrm{simulator}}$. Compute the new simulator model loss $\ell' {=} L\big(f_{\theta'_{\textrm{simulator}}}(x'), f_{\theta_{\textrm{task}}}(x') \big)$.
    \item Use the difference between $\ell$ and $\ell'$ as the proxy measure of explainability. The intuition is that the difference estimates how much extra information the explanation provides to the simulator model to forecast the task model's prediction.
\end{enumerate}

\paragraph{Details}
We repeat the experiments over $20$ test data-points ($10$ when the prediction is correct and incorrect), and choose five types of data for fine-tuning: random (with label neutral, entailment, or contradiction), most negative influential, and most positive influential. For a given test data-point and fine-tuning type, we fine-tune on $1$ data-point with $50$ learning rates in log-space from $10^{-5}$ to $10^{-2.5}$, and repeat for $10$ different fine-tuning data-points. The max and mins losses among the $10$ fine-tuning data-points are used to construct the confidence intervals.
Further, during the fine-tuning, the label we used corresponds to the true label (instead of the label predicted by the model).\footnote{We experimented with both settings in some initial experiments and found using the original/true labels performed better. This makes sense as the task model was trained with true label as targets, and fitting to the original labels replicates the process that produced the task model.
}

\paragraph{Extended Results}
Please see Fig.~\ref{appendix-figure:task-model-simulator-model-extended}.

\subsection{Effect Visualization}
\label{appendix-subsec:effect-visualization}
\paragraph{Experiment Design.}
We conduct experiments on two models, one trained on MultiNLI and the other trained on the HANS dataset. We then compute influence functions on their corresponding training data-points and build circular bipartite graphs. The nodes represent training (inner circle) and evaluation (outer circles) data-points (incorrect predictions), and the strength and color of edges represent the influence values. For visualization purposes, we organize the nodes so that positions of training data-points are determined via optimizing a weighted distance function to their connected test data-points. In the setting with the model trained on HANS, we further partition the training data-points (represented by three inner circles) based on the subset they are in. Note that, this is possible only because the influence function calculations are fast enough.

\paragraph{Details.}
In both settings (i.e., visualizations corresponding to the model trained on MultiNLI and HANS dataset), we select $400$ test data-points where the model predictions are incorrect, including $100$ from the MNLI evaluation dataset and $100$ for each of the three subsets from the HANS evaluation dataset. We use $k$NN with $k=10^3$.

\paragraph{Details on Computing Correlations.}
We slightly abuse the notation and define $\overline{\mathcal{I}}_{i, j}$ as the average (signed) influence values between the $i^{th}$ training data-point and $j^{th}$ data slice. Note that each training data-point can influence multiple evaluation data-points. We then compute the correlation between each of the two pairs of data slices $\rho(\overline{\mathcal{I}}_{\cdot, j_1}, \overline{\mathcal{I}}_{\cdot, j_2})$. We only include cases where the data-point has influences on both data slices.

\subsection{Error Correction}
\label{appendix-subsec:improving-model-prediction}

\paragraph{Experiment Design.}
\begin{enumerate}[leftmargin=*, noitemsep]
    \item For a given test dataset/data-slice, first evaluate the model performance on the data slice.
    \item Next, select a small batch of ``anchor'' data-points from the validation dataset/data-slice.
    \item Then, find influential data-point(s) from the training dataset w.r.t. those anchor data-points using influence functions.
    \item Update the model parameters by taking gradient step(s) on these influential data-point(s).
    \item Repeat Step~2-4 multiple times.
    \item Re-evaluate the model performance on the test dataset/data-slice.
\end{enumerate}

\paragraph{Why Fine-tuning?}
In the ideal scenario, one intuitive way to correct model predictions given influential examples is to retrain the model with these examples added to the training dataset. However, this is expensive as one could imagine.

To see why fine-tuning on influential examples makes sense, recall the definition of influence functions on model parameters,
\begin{equation}
\left.\mathcal{I}_{\text{params}}(z) := \frac{d \hat{\theta}_{\epsilon, z}}{d \epsilon}\right|_{\epsilon=0} \approx -H_{\hat{\theta}}^{-1} \nabla_{\theta} L(z, \hat{\theta})
\end{equation}
We can see that by setting the (matrix) learning rate proportional to $-H_{\hat{\theta}}^{-1}$, fine-tuning on influential examples is an approximation of retraining the model with the loss of influential examples upweighted. In practice, computing the Hessian inverse for each gradient update is expensive, so we could use a scalar learning rate instead (e.g., through diagonal approximation). This reduces to the vanilla fine-tuning setting. 

\paragraph{Details.}
The experiment is repeated $3$ times, and we compare the performance between using helpful data-points, harmful data-points, and random data-points. Since we assume gradient access to the anchor data-points, we further experiment with directly fine-tuning on those anchor data-points.

We mostly follow standard train/validation/test splits for different datasets. For HANS, we set aside $1\%$ of the evaluation dataset as the validation split, and for Amazon-WILDS, we use the OOD validation/test splits.
When the models are evaluated on HANS, we use each of the three slices of the HANS dataset as the evaluation dataset.

We repeat the Step~2-4 for $10$ iterations. For each iteration, we sample a batch size of $10$ from the validation dataset (i.e., the ``anchor'' data-points) when computing influence scores, and update model parameters for one gradient step on $10$ fine-tuning data-point with learning rate $10^{-4}$. For Amazon-WILDS experiments, we use $50$ anchor and fine-tuning data-points instead.

\paragraph{Discussion on~\citet{yang2020g}}
Recently,~\citet{yang2020g} examined the application of influence functions in filtering out detrimental synthetic training examples. Here we use them to select helpful training examples from another non-synthetic dataset. Both demonstrate that computing influence values on new examples could be useful. 
However,~\citet{yang2020g} (Appendix C) reported that with a pool of more than $380$K candidates, running influence functions could take more than $8$ hours. In our setup (i.e., finding a handful of influential examples), we can find the most helpful examples significantly faster thanks to our fast influence functions (see Main Paper Sec.~\ref{subsec:computation-times}). Extending our fast influence functions to their setup would be an interesting next-step.

\paragraph{Validation Results}
Please see Figs.~\ref{appendix-figure:data-augmentation-eval-ab} and~\ref{appendix-figure:data-augmentation-eval-cd}.

\end{document}